\documentclass[10pt,twocolumn,letterpaper]{article}

\usepackage{cvpr}              
\usepackage{multirow}
\usepackage{makecell}
\usepackage[most]{tcolorbox}
\usepackage{listings}
\tcbuselibrary{listings, breakable, skins}
\usepackage{setspace}
\usepackage{subcaption}

\usepackage{booktabs, makecell, multirow, xcolor, colortbl}
\usepackage{pifont}
\usepackage{bm}
\usepackage{graphicx}
\usepackage{tikz}
\usetikzlibrary{positioning,fit,calc,backgrounds}

\definecolor{cvprblue}{rgb}{0.21,0.49,0.74}
\usepackage[pagebackref,breaklinks,colorlinks,allcolors=cvprblue]{hyperref}

\def\confName{CVPR}
\def\confYear{2026}

\title{
\raisebox{-0.2\height}{\includegraphics[height=2em]{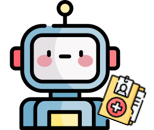}}\hspace{0.3em}
MedFG-VQA: Low-Frequency Memory and Graph Attention for Lightweight Medical VQA}

\author{
Haowen Gu$^{1,2}$, 
Gensheng Pei$^{3}$, 
Zeren Sun$^{1,2}$, 
Mingwu Ren$^{1,2}$\thanks{Corresponding author.}, 
Xiangbo Shu$^{1}$, 
Yazhou Yao$^{1,2}$\footnotemark[1], 
Fumin Shen$^{4}$\\
\small{$^{1}$School of Computer Science and Engineering, Nanjing University of Science and Technology}\\
\small{$^{2}$State Key Laboratory of Intelligent Manufacturing of Advanced Construction Machinery} \\
\small{$^{3}$Department of Electrical and Computer Engineering, Sungkyunkwan University} \\
\small{$^{4}$School of Computer Science and Engineering, University of Electronic Science and Technology of China} \\
\small{\url{https://github.com/NUST-Machine-Intelligence-Laboratory/MedFG}}
}

\begin{document}
\maketitle
\begin{abstract}
Medical Visual Question Answering (Med-VQA) holds significant promise for clinical decision support, yet faces challenges due to limited annotated data and the high computational demands of existing large vision-language 
models. We propose MedFG-VQA, a lightweight framework that leverages a memory bank to augment DCT-based low-frequency features and employs graph-enhanced cross-attention for effective visual-textual alignment. Specifically, our approach features two key components: Frequency-Memory Fusion (FMF), which enhances low-frequency features by retrieving from a learnable memory bank built on DCT decomposition, and Graph-Aware Cross-Attention (GACA), which aligns visual-textual features via cross-attention and refines them through graph-convolutional aggregation. To address data scarcity, we construct SynMed-VQA, a large-scale synthetic dataset comprising over 2 million question-answer pairs across 9 imaging modalities and 10 major organs, generated with GPT-4o. Extensive experiments on SynMed-VQA and three other standard biomedical VQA benchmarks demonstrate that MedFG-VQA achieves competitive or superior performance compared to much larger models while maintaining significantly lower computational costs, highlighting its efficiency and potential for clinical deployment.
\end{abstract}    
\vspace{0.2cm}
\section{Introduction}
\label{sec:intro}

\begin{figure}[t]
  \centering
   \includegraphics[width=\linewidth]{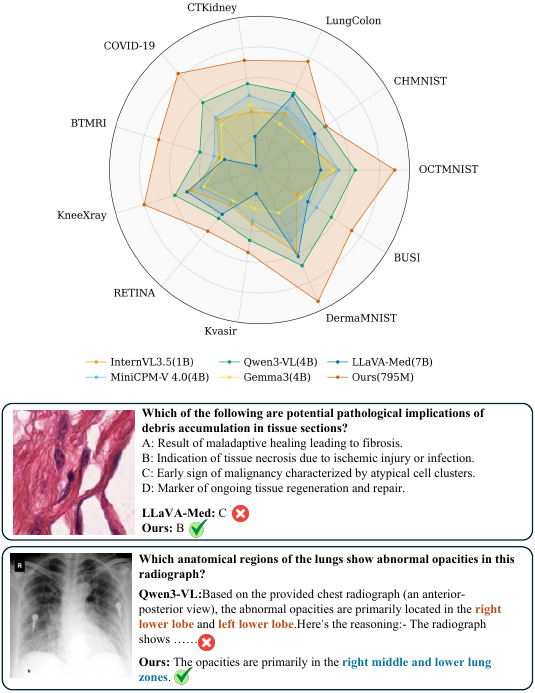}
   \caption{Illustration of VLMs evaluated on the SynMedVQA dataset. Among the 6 VLMs, MedFG-VQA achieves the highest overall score. Qualitative VQA comparison between two models, showcasing the effectiveness of MedFG-VQA.}
   \label{fig:figure1}
   \vspace{-0.6cm}
\end{figure}

Radiological imaging serves as a cornerstone of modern medicine, producing over 80 million images each year ~\cite{akhter2023ai}. With the growing demand for diagnostic interpretation, Medical Visual Question Answering (VQA) has emerged as a promising direction with substantial clinical relevance. By linking visual content in medical images with natural language queries, it enables more effective diagnostic assistance, image retrieval, and clinical decision support.

In recent years, the rapid advancement of Large Language Models (LLMs) and Vision-Language Models (VLMs) ~\cite{achiam2023gpt,yang2025qwen3,team2025gemma,zhu2023minigpt} has brought new opportunities for the development of medical VQA. However, unlike general-domain VQA, medical VQA faces two major challenges. First, The lack of high-quality annotated data, especially for cross-modal tasks requiring domain-specific medical knowledge. Although several public datasets have been released~\cite{liu2021slake, he2020pathvqa, lau2018dataset,liu2025gemex,chen2025mimo}, their limited scale makes them insufficient for training large VLMs. Second, clinical deployment imposes strict constraints on model size and computational resources, while existing models~\cite{chen2024huatuogpt, li2023llava,gai20253dradcomprehensive3dradiology,jiang2025omnivmedscalingmedicalvisionlanguage,lai2025medr1reinforcementlearninggeneralizable,nath2025vilam3enhancingvisionlanguagemodels} fail to maintain strong diagnostic capability under lightweight configurations.

Motivated by the aforementioned gap, we propose MedFG-VQA, a lightweight medical visual question answering model with frequency graph fusion, achieving efficient learning and strong generalization through structured module design and high-quality synthetic data. First of all, MedFG-VQA employs a pretrained visual backbone and introduces a lightweight FreqMemoryFusion (FMF) module. By retrieving and residually integrating low-frequency priors in the frequency domain, FMF enhances the model's capacity to capture global structural information. Then, we design a Graph-Aware Cross-Attention (GACA) module to jointly model global cross-modal semantics and local visual structure. Given image and text features, GACA achieves global cross-modal alignment through multi-head mutual attention, producing semantically enriched image representations. Meanwhile, a dynamic KNN-based graph convolution captures local spatial relationships among image patches. A gated residual fusion mechanism then adaptively balances these two complementary perspectives, and the resulting multimodal features are fed into an LLM to accomplish the VQA task. Equipped with the above methods, MedFG-VQA seamlessly integrates frequency-domain global modeling and graph-based local structural reasoning. To train MedFG-VQA, we construct a large-scale medical VQA synthetic dataset SynMedVQA comprising 2.059 million samples, generated with the assistance of GPT-4o. The dataset covers diverse medical scenarios and question types, enabling comprehensive model training. As shown in \ref{fig:figure1}, extensive experiments on multiple benchmarks demonstrate that MedFG-VQA achieves competitive performance with significantly reduced model size and computational cost.

Our contributions are as follows:

(1) We introduce SynMedVQA, a large-scale synthetic multimodal dataset generated via GPT-4o, comprising 2.059 million Q\&A pairs. The dataset spans 9 imaging modalities across 10 major organs, offering diverse and comprehensive supervision for medical VQA tasks.

(2) We develop FreqMemoryFusion (FMF), a novel module that leverages a learnable, frequency-domain memory bank. FMF retrieves low-frequency components and injects global structural priors through residual fusion, enhancing the robustness and generalization of lightweight models on structurally-oriented medical questions.

(3) We present Graph-Aware Cross-Attention (GACA), which combines cross-modal attention with a feature-adaptive KNN-based GCN. Adaptively fusing global semantic and local topological information via a gated mechanism, GACA improves alignment between fine-grained visual features and textual descriptions.

(4) Extensive experiments and ablation studies on multiple medical VQA benchmarks demonstrate that our approach achieves competitive performance. Notably, it does so using significantly fewer parameters than mainstream large models, validating the feasibility of small vision language models (SVLMs) in clinically relevant scenarios.
\section{Related Work}
\label{sec:related}
\textbf{Vision Language Model.}
With the rapid advancement of LLMs and the advent of large-scale pre-trained visual models like CLIP~\cite{radford2021learning}, numerous vision language models~\cite{li2022blipbootstrappinglanguageimagepretraining,zhang2025biomedclipmultimodalbiomedicalfoundation} have been developed to align image features with LLMs for comprehensive visual understanding. Recently, autoregressive architectures~\cite{li2023blip2bootstrappinglanguageimagepretraining,liu2023visualinstructiontuning} have gained popularity in the VLM domain, where many approaches feed both image features and textual inputs into LLMs to perform vision language tasks. However, these methods often depend on large visual encoders or complex feature alignment modules, resulting in high parameter counts and computational overhead, which limits their deployment in resource-constrained environments. Some studies~\cite{rang2025eve, diao2025evev2} have explored removing the visual encoder completely, directly entering raw image patches along with text into the LLM. Although this simplifies the model architecture, it can neglect local structural and spatial information, making it difficult to achieve strong performance on fine-grained visual understanding tasks.

\begin{figure*}[t]
  \centering
   \includegraphics[width=\linewidth]{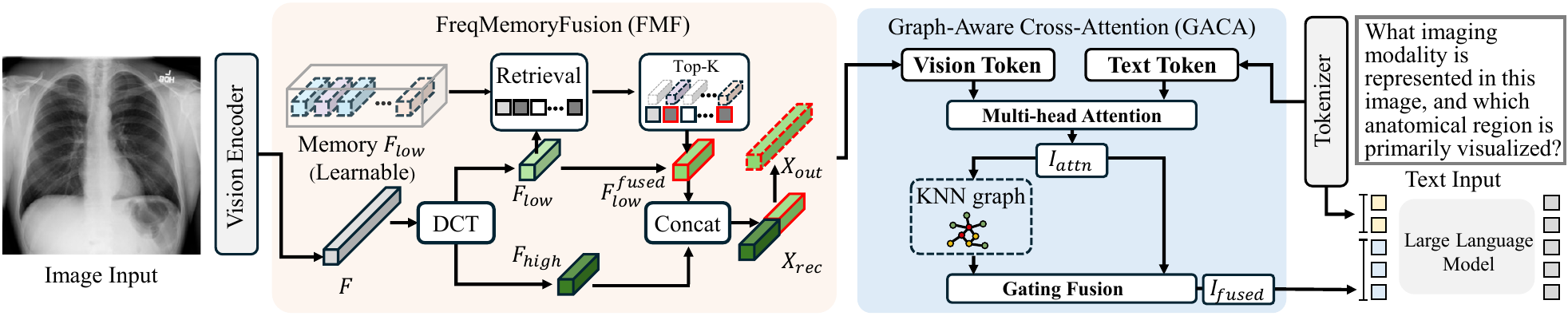}
   \caption{The overall architecture of MedFG-VQA. The model consists of a vision encoder, a FreqMemoryFusion(FMF) module, a Graph-Aware Cross-Attention(GACA) module and a LLM.}
   \label{fig:network}
   \vspace{-1em}
\end{figure*}

\noindent\textbf{Vision Language Model for Medical VQA.}
Driven by the development of large-scale vision language models, research in medical VQA has also made significant progress. BiomedCoOp~\cite{koleilat2025biomedcoop} leverages LLMs to achieve semantic consistency through prompt ensembling and combines this with a statistic-based prompt selection strategy for knowledge distillation, enabling efficient learning of prompt contexts. LLaVA-Med~\cite{li2023llava}, on the other hand, fine-tunes general VLMs on GPT-4 generated synthetic data, demonstrating remarkable performance on medical multimodal dialogue tasks. More recently, models such as LLADA-MedV~\cite{dong2025llada} employ diffusion-based vision language models and adopt visual instruction tuning to enhance understanding of biomedical images. Despite these advances, existing methods still face challenges in model lightweighting and fine-grained cross-modal feature alignment, limiting their applicability in resource-constrained clinical environments.

\noindent\textbf{Small Vision Language Model.} 
Recently, small or lightweight vision language models (sVLMs) have emerged to address the efficiency and scalability challenges of large VLMs. Models such as SmolVLM~\cite{marafioti2025smolvlmredefiningsmallefficient}, MiniGPT~\cite{zhu2023minigpt}, and MobileVLM~\cite{chu2023mobilevlmfaststrong} employ compact visual backbones or low-rank adaptation strategies to substantially reduce computational overhead while preserving strong multimodal reasoning capabilities. However, most existing research has focused on developing general-purpose small models and improving performance on broad-domain benchmarks, with comparatively limited efforts devoted to domain-specific or specialized applications such as medical imaging. This gap can be partly attributed to the fact that small models are more dependent on high-quality, domain-relevant training data. Recent work such as DataRater~\cite{calian2025dataratermetalearneddatasetcuration} has shown that training with a smaller amount of high-quality data can outperform training on large but noisy datasets, highlighting the crucial role of data quality in small-model performance. In the medical domain, however, the limited availability of carefully curated datasets and privacy constraints further hinder the development of lightweight VLMs tailored for tasks like medical visual question answering.

\section{Method}

\label{sec:method}
An overview of the proposed MedFG-VQA framework is illustrated in Figure \ref{fig:network}. By leveraging a pre-trained vision encoder to extract rich visual features, our method introduces two key components for effective multimodal feature learning. Specifically, the FMF module enhances low-frequency representations by retrieving from a learnable memory bank and residual integration, while the GACA module aligns visual-textual features via cross-attention and aggregates local spatial context through KNN-based graph convolution. The fused multimodal features are fed into a large language model, generating the final answer and enabling accurate, robust performance on medical VQA tasks.
\subsection{FreqMemoryFusion}
\label{subsec:FMF}
To better exploit the frequency-domain structural properties of visual representations and enhance the model's capacity for global low-frequency modeling, we introduce the Frequency Memory Fusion (FMF) module. Given an input feature $\mathbf{X} \in \mathbb{R}^{B \times M \times D}$, we first transform it into the frequency domain via Discrete Cosine Transform (DCT), decomposing it into low-frequency $\mathbf{F}_\text{low}$ and high-frequency $\mathbf{F}_\text{high}$ components. Since low-frequency components predominantly encode global semantic and structural information of the image, we leverage them as query signals for global prior retrieval.

Specifically, we maintain a learnable memory bank and retrieve the top-k memory entries $\mathbf{M}_k$ and corresponding similarity weights $\mathbf{S}_k$ based on cosine similarity with the current low-frequency feature $\mathbf{F_\text{low}}$. A residual weighted fusion strategy then integrates these entries, yielding the enhanced low-frequency representation:
 
\begin{equation}
  \mathbf{F}_{\mathrm{low}}^{\mathrm{fused}} = \lambda \mathbf { F } _ {\text{low}} + ( 1 - \lambda )(\text{Softmax} ( \mathbf{S}_k ) \cdot \mathbf{M}_k),
  \label{eq:F_low_fuse}
\end{equation}
where $\lambda \in [0,1]$ is a fusion coefficient that balances the contribution between the original feature and the memory-enhanced feature. This design enables the model to incorporate global structural priors while maintaining feature detail fidelity and representation stability. In practice, we fix $\lambda = 0.7$ based on preliminary experiments, which achieves a good trade-off between feature fusion and stability.

Subsequently, the enhanced low-frequency features $\mathbf{F}_{\mathrm{low}}^{\mathrm{fused}}$ are concatenated with the original high-frequency components $\mathbf{F_{high}}$ and transformed back to the spatial domain via Inverse Discrete Cosine Transform (IDCT), producing the reconstructed feature $\mathbf{X}_{rec}$. To adaptively integrate the reconstructed and original features, we employ a lightweight gated residual fusion mechanism:
 
\begin{equation}
  \mathbf{X}_{\text{out}} = \mathbf{X} + \alpha \cdot f_{\theta}([\mathbf{X}, \mathbf{X}_{\text{rec}}]),
  \label{eq:X_out}
\end{equation}
where $\alpha$ is a learnable gating parameter and $f_{\theta}(\cdot)$ is a linear projection that adaptively fuses the enhanced features with the original input for effective feature refinement.

\noindent\textbf{Learnable Memory.} In FMF, a learnable global memory $\mathbf{M} \in \mathbb{R}^{N \times \frac{D}{2}}$ is introduced to preserve global low-frequency representations, where $N$ denotes the number of memory entries. The memory vectors are orthogonally initialized before training and continuously updated during optimization, allowing the memory to gradually learn representative global priors that capture stable low-frequency feature patterns inherent in the data distribution. However, without proper regularization, the learned memories may collapse into redundant or overlapping representations. To address this, we introduce a diversity loss that encourages the memory embeddings to remain distinct and informative:
\begin{equation}
    { \mathcal{L} _ \text{ div }} = { \frac { 1 } { N ( N - 1 ) } } \sum _ { i \neq j } \left( \mathbf { m } _ { i } ^ { \mathsf { T } } \mathbf { m } _ { j } \right) ^ { 2 }.
    \label{L_div}
\end{equation}
Here, $\mathbf{m}_i$ denotes the $i_{th}$ vector in the memory bank. By minimizing the off-diagonal similarities, the loss encourages the memory vectors to remain diverse, thereby enhancing the effectiveness of feature fusion.
\subsection{Graph-Aware Cross-Attention}
In Graph-Aware Cross-Attention(GACA) module, we design both global semantic interactions across modalities and local structural modeling of visual features. Given image features $\mathbf{I} \in \mathbb{R}^{B \times M \times D}$ and text features $\mathbf{T} \in \mathbb{R}^{B \times T \times D}$, we first perform layer normalization on both modalities. We then treat the image features as queries while the text features serve as keys and values. Through multi-head cross-attention, we achieve global cross-modal alignment and obtain semantically enriched visual representations $\mathbf{I}_{\mathrm{attn}}$.

To further incorporate local spatial relationships, we construct a dynamic KNN graph based on $\mathbf{I}_{\mathrm{attn}}$, where each node represents an image patch and edges form according to feature similarity. The adjacency matrix is defined as:

\begin{equation} 
A _ { i j } = { \left\{ \begin{array} { l l } { 1 , } & { { \mathrm { i f ~ } } j \in \mathrm { K N N } ( i , k ) , } \\ { 0 , } & { { \mathrm {otherwise.} } } \end{array} \right. } \label{A_ij} 
\end{equation}

After symmetrizing and normalizing the adjacency matrix to obtain $\tilde{\mathbf{A}}$, we apply a graph convolutional layer to propagate and aggregate information from neighboring nodes, thereby enhancing contextual coherence among locally related visual regions:
\begin{equation}
\mathbf{I}_{\text{enh}} = \sigma(f_{\theta}(\tilde{\mathbf{A}} \mathbf{I}_{\text{attn}})),
\label{I_gcn}
\end{equation}
where $f_{\theta}$ denotes a learnable weight matrix and $\sigma$ is a non-linear activation function

Finally, a gated residual fusion mechanism is employed to adaptively integrate the cross-modal semantic representation $\mathbf{I}_{\text{attn}}$ and locally aggregated structural features $\mathbf{I}_{\text{enh}}$. Specifically, the gate value is computed from the concatenation of the two feature types, controlling their relative contributions in fusion. The final representation is obtained as:
\begin{equation}
    \mathbf{I}_{\text {fused }}=\mathbf{G} \odot \mathbf{I}_{\text {enh }}+(1-\mathbf{G}) \odot \mathbf{I}_{\text {attn }},
    \label{I_fuse}
\end{equation}
where $\mathbf{G} = \sigma(f_{\theta}([\mathbf{I}_{\text{attn}}, \mathbf{I}_{\text{enh}}]))$ denotes the learned gating weights, which are generated by a learnable linear projection followed by a sigmoid activation.
This design enables the model to dynamically balance global semantic alignment and local structural aggregation, yielding a multimodal representation that jointly captures both global and local visual information.

\begin{table}[t]
\centering
\caption{Overview of datasets statistics, including image and question counts for training, validation, and test splits. Open and Close questions share the same quantities.}
\vspace{-0.3cm}
\resizebox{\linewidth}{!}{
\begin{tabular}{lcc}
\toprule
\textbf{Dataset} & \textbf{\makecell{Images\\(train $|$ val $|$ test)}} & \textbf{\makecell{Questions\\(train $|$ val $|$ test)}} \\
\midrule
CHMNIST~\cite{kather2016multi} & 2496 $|$ 1000 $|$ 1504 & 12480 $|$ 5000 $|$ 7520 \\
RETINA~\cite{kohler2013automatic} & 2108 $|$ 841 $|$ 1268 & 10540 $|$ 4205 $|$ 6340 \\
BTMRI~\cite{msoud_nickparvar_2021} & 2854 $|$ 1141 $|$ 1717 & 14270 $|$ 5705 $|$ 8585 \\
DermaMNIST~\cite{tschandl2018ham10000} & 7006 $|$ 1003 $|$ 2005 & 35030 $|$ 5015 $|$ 10025 \\
BUSI~\cite{al2020dataset} & 389 $|$ 155 $|$ 236 & 1945 $|$ 775 $|$ 1180 \\
OCTMNIST~\cite{kermany2018identifying} & 97476 $|$ 10832 $|$ 1000 & 487380 $|$ 54160 $|$ 5000 \\
KneeXray~\cite{chen2018knee} & 5778 $|$ 826 $|$ 1656 & 28890 $|$ 4130 $|$ 8280 \\
COVID-19~\cite{tahir2021covid} & 10582 $|$ 4232 $|$ 6351 & 52910 $|$ 21160 $|$ 31755 \\
Kvasir~\cite{pogorelov2017kvasir} & 2000 $|$ 800 $|$ 1200 & 10000 $|$ 4000 $|$ 6000 \\
CTKidney~\cite{islam2022vision} & 6221 $|$ 2487 $|$ 3738 & 31105 $|$ 12435 $|$ 18690 \\
LungColon~\cite{borkowski2019lung} & 12500 $|$ 5000 $|$ 7500 & 62500 $|$ 25000 $|$ 37500 \\
\bottomrule
\end{tabular}
}
\label{tab:dataset_stats}
\vspace{-0.3cm}
\end{table}

\subsection{Training Strategy \& Loss Function}
For text generation, we employ the cross-entropy loss $\mathcal{L}_{\text{text}}$. To encourage the memory vectors in the FMF module to remain diverse, we compute the diversity loss $\mathcal{L}_{\text{div}}$ as described in \S\ref{subsec:FMF}. The final training objective is a linear combination of these losses, which can be formulated as:
\begin{equation}
    \mathcal{L}_\text{total} = \mathcal{L} _ \text{text} + \lambda \mathcal{L} _ \text{div}.
    \label{loss}
\end{equation}

\section{Dataset}

\label{sec:dataset}
\subsection{Data Source}

\begin{figure*}[t]
  \centering
   \includegraphics[width=\linewidth]{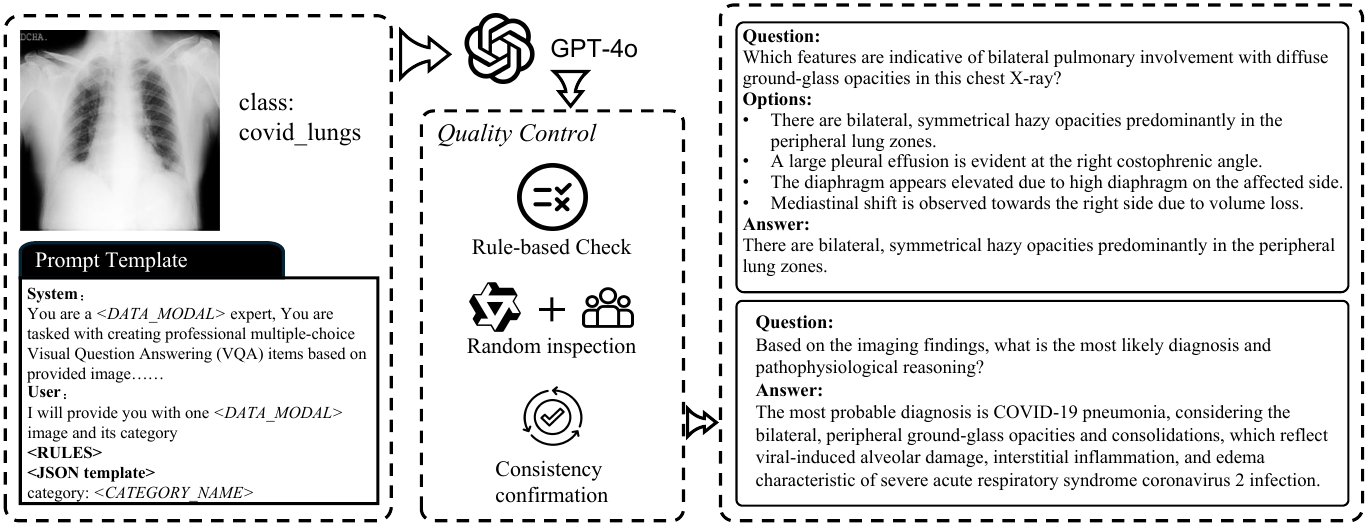}
   \vspace{-0.5cm}
   \caption{The construction pipeline of SynMedVQA. The pipeline includes prompt template, samples of Q\&A pair generated by GPT-4o and the quality control process.}
   \label{fig:dataConstruction}
   \vspace{-0.2cm}
\end{figure*}

We integrate 11 publicly available medical imaging datasets, \ie, BTMRI~\cite{msoud_nickparvar_2021}, BUSI~\cite{al2020dataset}, CHMNIST~\cite{kather2016multi}, COVID-19~\cite{tahir2021covid}, CTKidney~\cite{islam2022vision}, DermaMNIST~\cite{tschandl2018ham10000}, KneeXray~\cite{chen2018knee}, Kvasir~\cite{pogorelov2017kvasir}, LungColon~\cite{borkowski2019lung}, OCTMNIST~\cite{kermany2018identifying}, and RETINA~\cite{kohler2013automatic}.
These datasets collectively span nine distinct imaging modalities and ten major anatomical regions, offering a comprehensive and diverse benchmark for multimodal medical visual understanding. Specifically, the included imaging modalities encompass magnetic resonance imaging (MRI), computed tomography (CT), X-ray, ultrasound, dermoscopy, histopathology, fundus photography, optical coherence tomography (OCT), and endoscopy.
The datasets further cover a wide range of clinical targets and organs, including the brain, breast, lungs, liver and kidneys, retina, knee, and skin, providing both organ-level and cellular-level visual representations. Such a broad coverage enables our model to learn cross-modality feature alignment and generalizable reasoning patterns across heterogeneous medical domains.
By consolidating these diverse datasets, we establish a unified and large-scale foundation that enables comprehensive evaluation and robust training of vision language models for Med-VQA.
In total, the integrated dataset contains 205,902 medical images, divided into 149,410 for training, 28,317 for validation, and 28,175 for testing. The data splits strictly follow the configurations of BiomedCoOp~\cite{koleilat2025biomedcoop} to ensure comparability and consistency across benchmarks.

\subsection{Data Generation}
\label{sec:data_generation}
To support medical-scene-oriented multimodal visual question answering tasks, we constructed the \textbf{SynMedVQA} dataset by leveraging cleaned and structured image-label information and employing GPT-4o as a semantic generation engine to automatically produce \textit{Q\&A} pairs, the overall pipeline is shown as Figure \ref{fig:dataConstruction}. We develop a unified and adaptable structured prompting framework that casts the model as a senior medical imaging expert and dynamically tailors question generation to the clinical and radiological characteristics of each dataset. This design ensures domain-specific relevance while maintaining structural consistency across datasets. All generated questions follow a consistent schema targeting one of four core aspects: (1) imaging features (\eg, signal intensity, enhancement pattern, morphology), (2) visible anatomical structures, (3) pathological manifestations (\eg, mass effect, edema, invasion), and (4) clinical implications (\eg, symptoms, functional impact, management considerations). Within each aspect, emphasis is customized per disease domain, for instance, brain tumor questions prioritize contrast enhancement and midline shift, pulmonary nodule questions focus on margin characteristics and mediastinal involvement, and liver lesion questions highlight segmental localization and vascular invasion.

\begin{figure}[t]
\centering
\includegraphics[width=0.8\linewidth]{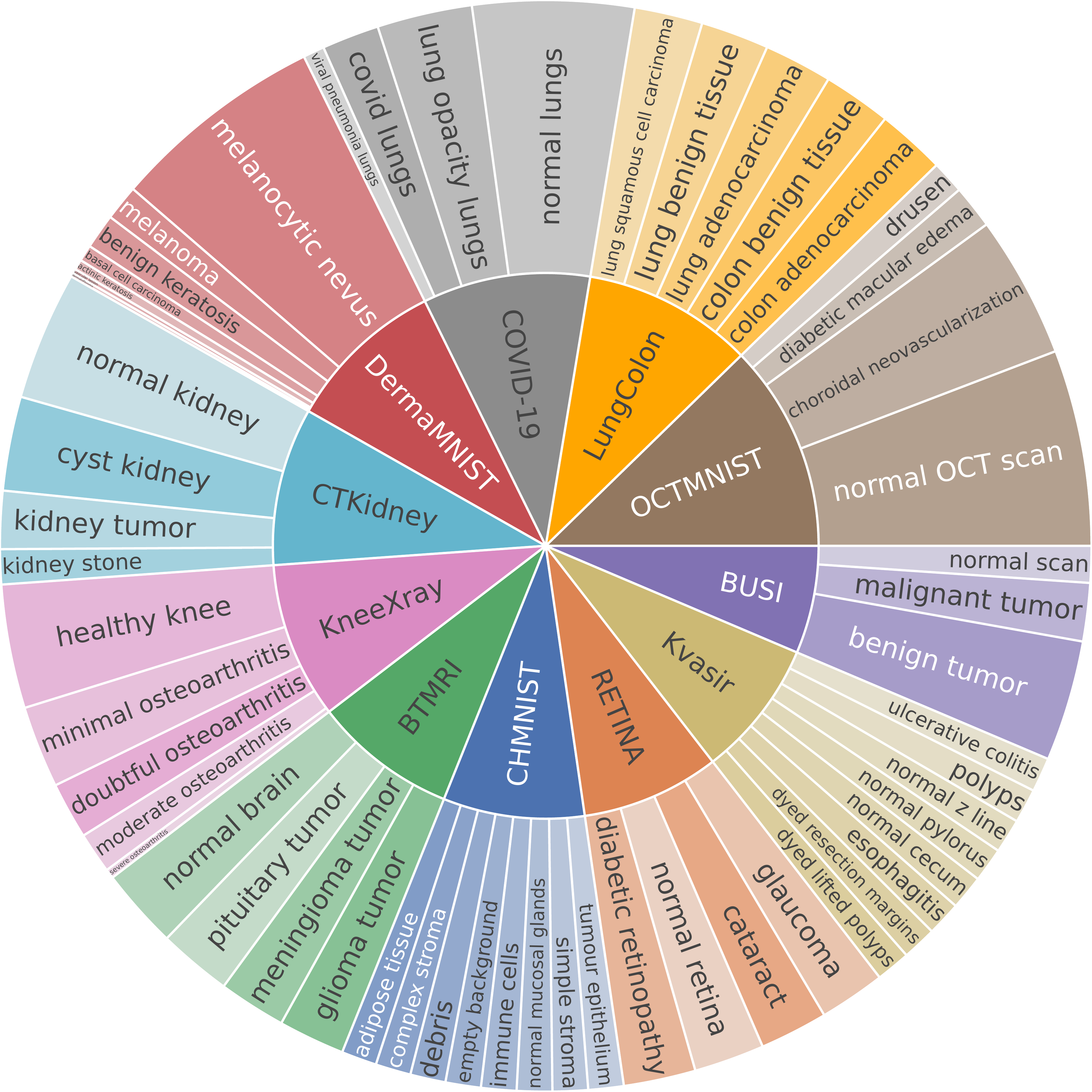}
\caption{Sunburst visualization of dataset category proportions, where the inner circle corresponds to datasets and the outer circle to category ratios within each dataset.}
\label{fig:dataset_overview}
\vspace{-0.6cm}
\end{figure}

For each image, we automatically generate two types of questions via GPT-4o. Open-ended questions include queries such as ``\textit{Which organ contains the lesion marked in the image?}" and ``\textit{What is the primary abnormality shown in this image?}". For multiple-choice questions, 75\% are designed with one correct answer and three distractors, while the remaining 25\% contain four distractors without a correct answer, enhancing both discriminability and robustness.

To ensure clinical accuracy and interpretability, prompt templates are integrated with original annotations, with multiple customized variants designed to address the diversity of imaging modalities and organ-specific characteristics across datasets. Each prompt is structured into System and User components: the former establishes the expert role, `` You are a senior medical imaging specialist in \textit{DATA\_MODAL} interpretation and \textit{DISEASE\_DOMAIN} diagnosis ", while the latter provides the image, lesion category, and specific task instructions. JSON formatting is enforced for all outputs to enable efficient rule-based validation.
Given the richness and diversity of these prompts, the complete designs and specifications are provided in the Appendix to facilitate reproducibility and future research.

\begin{table*}
\centering
\small
\caption{Comparison of model performance on SynMedVQA. All baseline models use their publicly available pretrained weights.}
\vspace{-0.2cm}
\resizebox{\linewidth}{!}{
    \renewcommand\arraystretch{1.2}
    \tabcolsep=1.2pt
    \begin{tabular}{lccccccccccccc}
    \toprule
    \textbf{Method} & 
    \textbf{Params} &
    \rotatebox{45}{OCTMNIST} &
    \rotatebox{45}{CHMNIST} &
    \rotatebox{45}{LungColon} &
    \rotatebox{45}{CTKidney} &
    \rotatebox{45}{COVID-19} &
    \rotatebox{45}{BTMRI} &
    \rotatebox{45}{KneeXray} &
    \rotatebox{45}{RETINA} &
    \rotatebox{45}{Kvasir} &
    \rotatebox{45}{DermaMNIST} &
    \rotatebox{45}{BUSI} &
    \textbf{\rotatebox{45}{Average}} \\
    \midrule
    InternVL3.5~\cite{wang2025internvl35advancingopensourcemultimodal}  & 1B & 0.538& 0.4569& 0.4802& 0.4714& 0.4943& 0.4261& 0.5147& 0.4355& 0.4588& 0.5662& 0.433 & 0.4846 \\
    MiniCPM-V 4.0~\cite{yao2024minicpm}  & 4B & 0.5374& 0.4964& 0.4987& 0.52& 0.503& 0.4425& 0.4838& 0.4276& 0.4503& 0.5288& 0.5033 & 0.4964 \\
    Qwen3-VL~\cite{yang2025qwen3}  & 4B & 0.5872 & 0.5334 & 0.5473 & 0.5547 & 0.5601 & 0.4869 & 0.5647 & 0.4877 & 0.5083 & 0.6076 & 0.5559 & 0.5492 \\
    Gemma3~\cite{team2025gemma}   & 4B & 0.5208 & 0.4537 & 0.4474 & 0.4927 & 0.4751 & 0.4162 & 0.4739 & 0.4202 & 0.4142 & 0.4387 & 0.4483 & 0.4590 \\
    LLaVA-Med~\cite{li2023llava} &7B & 0.4834 & 0.4961 & 0.5395 & 0.3991 & 0.3165 & 0.4096 & 0.5272 & 0.4719 & 0.3703 & 0.5788 & 0.4720 & 0.4834 \\
    \textbf{Ours}   &795M & \textbf{0.7056} & \textbf{0.5372} & \textbf{0.6493} & \textbf{0.6239} & \textbf{0.6733} & \textbf{0.6155} & \textbf{0.6609} & \textbf{0.5372} & \textbf{0.5440} & \textbf{0.7225} & \textbf{0.6284} & \textbf{0.6441} \\
    \bottomrule
    \end{tabular}
}
\label{tab:SynMedVQA}
\vspace{-0.4cm}
\end{table*}

To ensure high quality, the generated \textit{Q\&A} pairs are subjected to a rigorous three-step validation process. First, we apply automated, rule-based filtering to discard duplicate, ambiguous, or incomplete questions. Next, a random subset is reviewed using a combination of Qwen2.5-VL~\cite{bai2025qwen25vltechnicalreport} and manual human inspection. This step verifies that the questions are clear, complete, and relevant. In the final stage, we conduct a consistency check to ensure questions, answers, and original image annotations are logically aligned, eliminating any contradictions. This comprehensive process guarantees the dataset is high-quality and well-suited for training and evaluating medical VQA models.

The SynMedVQA dataset comprises 2,059,020 \textit{Q\&A} pairs, equally split between open-ended and multiple-choice questions. The questions cover multiple aspects, including anatomical structures, imaging characteristics, pathological changes, and clinical manifestations. This comprehensive scope provides a diverse and clinically grounded benchmark for advancing multimodal reasoning in medical vision language models.
Table~\ref{tab:dataset_stats} details the dataset's statistics, while Figure~\ref{fig:dataset_overview} illustrates the category distributions using a sunburst visualization. In this figure, the inner circle represents the datasets, and the outer circle shows the category proportions within each.

\section{Experiments}
\label{sec:exp}

\subsection{Experimental Setup}

\textbf{Datasets and Metrics.}
Our model is trained and validated on the SynMedVQA dataset memtioned at \S\ref{sec:dataset}.
Meanwhile, to further evaluate its generalization ability, we also conduct experiments on three public medical VQA benchmarks: SLAKE~\cite{liu2021slake}, VQA-RAD~\cite{lau2018dataset}, and PathVQA~\cite{he2020pathvqa}.
Following the evaluation protocol of LLaVA-Med~\cite{li2023llava}, we measure model performance using answer accuracy. For closed-ended questions, correctness is directly determined by comparing the model’s response with the ground truth. For open-ended questions, we evaluate model accuracy by constructing multiple-choice options. These options are formed by pairing the reference answer with several alternative responses sampled from the training set, and the model is assessed on its ability to select the correct one.

\noindent{\textbf{Implementation Details.}}
We adopt SigLIP2-so400m~\cite{tschannen2025siglip2multilingualvisionlanguage} as the visual backbone and freeze its parameters during training. To improve computational efficiency, modality projection (MP)~\cite{shi2016real} is applied to reduce the number of visual tokens. The large language model is SmolLM2-360M-Instruct~\cite{allal2025smollm2smolgoesbig}, along with its corresponding tokenizer. Both FMF and GACA modules are randomly initialized before training. After initialization, the entire model is trained on the proposed SynMedVQA dataset for 2 epochs. The learning rates are set as follows: 5e-5 for the LLM, 0.003 for the modality projector, and 0.0015 for the FMF and GACA modules. We use AdamW~\cite{loshchilov2019decoupledweightdecayregularization} as the optimizer, and all experiments are conducted on 8×4090 GPUs.

\begin{table*}[t]
\centering
\renewcommand{\arraystretch}{1.1}
\tabcolsep=7pt
\caption{Comparative results on three public medical VQA benchmarks. The best and second-best results are indicated in \textbf{bold} and \underline{underline}, respectively.}
\resizebox{0.93\linewidth}{!}{
\begin{tabular}{lccc|cc|cc|cc}
\toprule
\multirow{2}{*}{\textbf{Model}} & 
\multicolumn{3}{c|}{\textbf{Params}} &
\multicolumn{2}{c}{\textbf{SLAKE}~\cite{liu2021slake}} &
\multicolumn{2}{c}{\textbf{VQA-RAD}~\cite{lau2018dataset}} &
\multicolumn{2}{c}{\textbf{PathVQA}~\cite{he2020pathvqa}} \\
\cmidrule(lr){2-4} \cmidrule(lr){5-6} \cmidrule(lr){7-8} \cmidrule(lr){9-10}
 & \textbf{Vision} & \textbf{LLM} & \textbf{Total} & \textbf{Closed} & \textbf{Open} & \textbf{Closed} & \textbf{Open} & \textbf{Closed} & \textbf{Open} \\
\midrule
InternVL3.5~\cite{wang2025internvl35advancingopensourcemultimodal} & 0.3B & 0.8B & 1.1B & \underline{0.6459} & 0.8156 & 0.6140 & 0.6180 & 0.5694 & 0.3187 \\
MiniCPM-V 4.0~\cite{yao2024minicpm} & 0.4B & 3.0B & 4.1B & 0.4246 & 0.8315 & \underline{0.7353} & 0.7978 & 0.6376 & 0.4513 \\
Qwen3-VL~\cite{yang2025qwen3} & 0.4B & 4.0B & 4.4B & 0.5120 & \underline{0.9006} & \textbf{0.7868} & \textbf{0.8764} & 0.6679 & \underline{0.5249} \\
Gemma3~\cite{team2025gemma} & 0.4B & 0.6B & 3.2B & 0.5287 & 0.7337 & 0.6507 & \underline{0.8090} & \textbf{0.8410} & 0.4608 \\
LLaVA-Med~\cite{li2023llava} & 0.3B & 7.2B & 7.5B & \textbf{0.6567} & 0.1232 & 0.6434 & 0.0730 & 0.7308 & 0.2828 \\
\textbf{Ours} & 412M & 316M & 795M & 0.5502 & \textbf{0.9595} & 0.6324 & 0.7865 & \underline{0.6694} & \textbf{0.8062} \\
\bottomrule
\end{tabular}
}
\label{tab:opendatasets}
\end{table*}

\subsection{Comparison with Previous Studies}
We systematically evaluate model performance on the SynMedVQA benchmark and compare it with five representative vision language models: InternVL3.5~\cite{wang2025internvl35advancingopensourcemultimodal}, MiniCPM-V 4.0~\cite{yao2024minicpm}, Qwen3-VL~\cite{yang2025qwen3}, Gemma3~\cite{team2025gemma}, and LLaVA-Med v1.5~\cite{li2023llava}.
To ensure fair comparison, all models are evaluated under the same data splits and metric. Notably, the baseline models use their publicly available pretrained weights without additional fine-tuning on SynMedVQA.

As shown in Table \ref{tab:SynMedVQA}, our model achieves an average accuracy of 0.6441, surpassing next-best model Qwen3-VL by approximately 9.5\%, despite having only 795M parameters significantly fewer than all other models. Notably, the model demonstrates superior performance on tasks such as OCTMNIST, CTKidney, COVID-19, and BTMRI, which involve complex anatomical structures or significant lesion variations. These results suggest that the proposed frequency-domain enhancement and cross-modal graph structure modeling effectively improve multimodal understanding and diagnostic reasoning, even under a lightweight model design.

We further evaluate our model on three public medical VQA benchmarks: SLAKE~\cite{liu2021slake}, VQA-RAD~\cite{lau2018dataset}, and PathVQA~\cite{he2020pathvqa}, covering both closed-ended and open-ended question types. The results are summarized in Table~\ref{tab:opendatasets}.

Although our model contains only 795M parameters, which is significantly smaller than the other compared models, it achieves competitive performance across multiple benchmarks. In open-ended tasks, it attains the highest accuracy among all models, demonstrating effective generalization in reasoning over complex semantic questions. LLaVA-Med shows low performance on open-ended tasks because it does not follow the instructions to provide direct answer choices, making accuracy evaluation unsuitable. On closed-ended questions, some larger models perform slightly better, which can be attributed to their substantially larger pretraining datasets while our model only trained on SynMedVQA.

\begin{figure}[t]
\centering
\small
\captionof{table}{Ablation studies of the contribution of FMF and GACA, the effect of the memory bank size in FMF, and the impact of the loss balance coefficient $\lambda$.}
\label{tab:ablation}

\begin{subtable}[t]{0.32\linewidth}
    \centering
    \begin{tabular}{ccc}
        \toprule
        FMF & GACA & Acc. \\
        \midrule
        &  & 0.6270 \\
        \checkmark &  & 0.6242 \\
        & \checkmark & 0.4170 \\
        \checkmark & \checkmark & \textbf{0.6441} \\
        \bottomrule
    \end{tabular}
    \caption*{(a)} 
\end{subtable}%
\hfill 
\begin{subtable}[t]{0.15\linewidth}
    \centering
    \begin{tabular}{cc}
        \toprule
        Size & Acc. \\
        \midrule
        16 & 0.4313 \\
        32 & 0.6385 \\
        64 & \textbf{0.6441} \\
        128 & 0.2771 \\
        \bottomrule
    \end{tabular}
    \caption*{(b)} 
\end{subtable}%
\hfill 
\begin{subtable}[t]{0.25\linewidth}
    \centering
    \begin{tabular}{cc}
        \toprule
        $\lambda$ & Acc. \\
        \midrule
        0.3 & 0.6415 \\
        0.5 & \textbf{0.6441} \\
        0.7 & 0.6427 \\
        0.9 & 0.6410 \\
        \bottomrule
    \end{tabular}
    \caption*{(c)} 
\end{subtable}
\vspace{-0.6cm}
\end{figure}

\subsection{Ablation Studies}

To verify the effectiveness of the proposed Frequency-domain Memory Fusion (FMF) and Graph-based Attention Cross-modal Alignment (GACA) modules, we conducted a series of ablation experiments. The experimental results are presented in Table \ref{tab:ablation} and Table \ref{tab:ablation2}.

\begin{figure*}[t]
    \centering
    \begin{subfigure}{0.34\textwidth}
        \includegraphics[width=\linewidth]{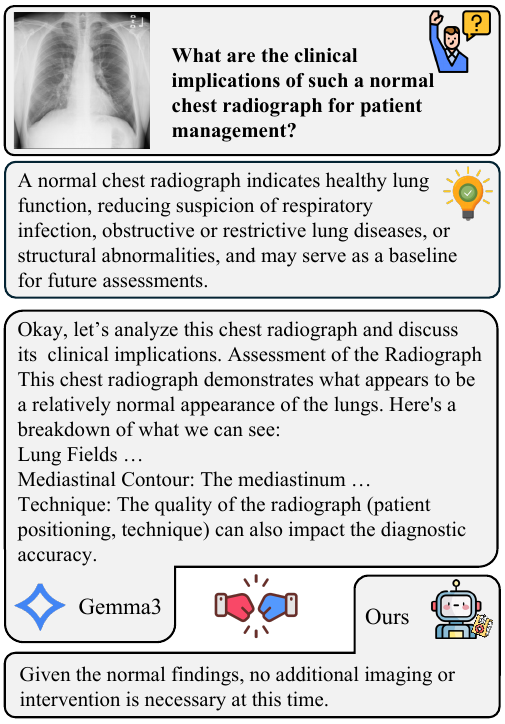}
        \caption{}
    \end{subfigure}
    \begin{subfigure}{0.32\textwidth}
        \includegraphics[width=\linewidth]{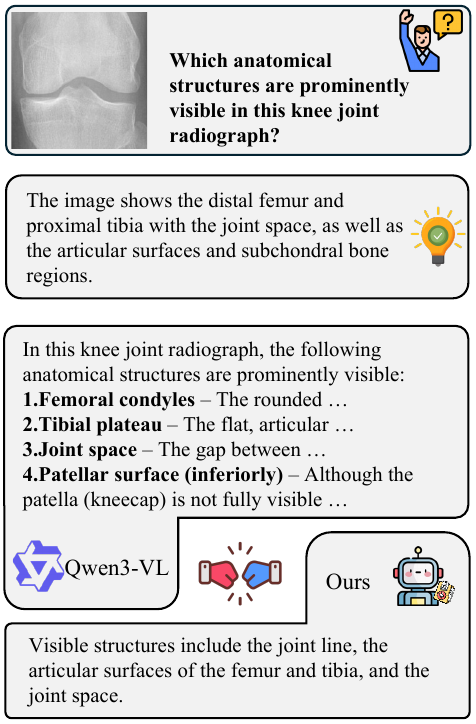}
        \caption{}
    \end{subfigure}
    \begin{subfigure}{0.32\textwidth}
        \includegraphics[width=\linewidth]{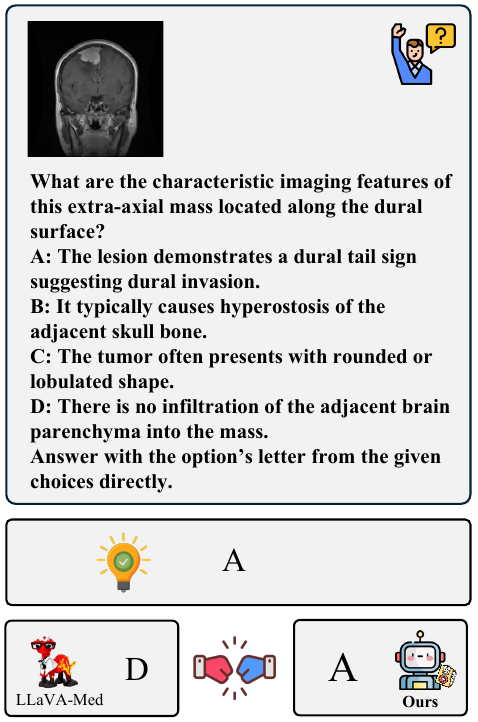}
        \caption{}
    \end{subfigure}
    \vspace{-0.2cm}
    \caption{Qualitative results against (a) Gemma3-4B , (b) Qwen3-VL-4B, and (c) LLaVA-Med v1.5. Responses are abridged for brevity.}
    \label{fig:qualitative}
    \vspace{-0.3cm}
\end{figure*}

It can be observed in Table~\ref{tab:ablation}(a) that removing both FMF and GACA modules leads to a significant performance drop, with accuracy falling to 0.627. The combined application of FMF and GACA leads to significant performance gains, effectively strengthening global feature modeling and cross-modal semantic alignment. When both modules are integrated, the accuracy rises to 0.6441, substantially outperforming the baseline. This demonstrates that FMF effectively strengthens global structural perception of visual features, while GACA enhances semantic interactions between images and text, and their synergistic combination leads to a marked improvement in overall performance.

\begin{table}[t] 
\centering
\small

\caption{Ablation studies on different frequency-domain transformation strategies and module against standard cross-attention.}
\label{tab:ablation2}

\begin{subtable}[t]{0.45\linewidth}
    \centering
    \begin{tabular}{cc}
        \toprule
        Type & Acc. \\
        \midrule
        -- & 0.6058 \\
        FFT & 0.5585 \\
        DCT & \textbf{0.6441} \\
        \bottomrule
    \end{tabular}
    \caption*{(a)} 
\end{subtable}
\hfill 
\begin{subtable}[t]{0.45\linewidth}
    \centering
    \begin{tabular}{cc}
        \toprule
        Method & Acc. \\
        \midrule
        - & - \\
        CA & 0.6407 \\
        GACA & \textbf{0.6441} \\
        \bottomrule
    \end{tabular}
    \caption*{(b)} 
\end{subtable}
\vspace{-0.5cm}
\end{table}

Table~\ref{tab:ablation}(b) shows the impact of memory bank size on model performance. Increasing the memory size from 16 to 64 notably improves accuracy, indicating that a larger memory helps capture richer contextual representations. However, further increasing it to 128 leads to a sharp drop, suggesting that excessive memory introduces redundant or noisy information that hinders retrieval. Thus, a memory bank size of 64 is adopted as a balanced choice.

Table~\ref{tab:ablation}(c) presents the effect of the loss balance factor $\lambda$ on model performance. As shown in the table, varying $\lambda$ from 0.3 to 0.9 results in only minor fluctuations in accuracy, with the best performance achieved at $\lambda=0.5$, reaching an accuracy of 0.6441. Combined with the findings from Table~\ref{tab:ablation}(a) and Table~\ref{tab:ablation}(b), this indicates that the model’s performance is relatively insensitive to the exact choice of $\lambda$. Nevertheless, the memory bank diversity loss remains essential, as it contributes to the richness of the representations captured in the memory bank and underpins the improvements observed in global feature modeling and cross-modal alignment.

Table~\ref{tab:ablation2}(a) analyzes the effectiveness of frequency-domain transformations in the FMF module. We compare directly feeding image features, applying Fast Fourier Transform (FFT), and applying Discrete Cosine Transform (DCT). FFT achieves the best performance with an accuracy of 0.6441, outperforming both the direct and DCT-based settings. This demonstrates that FFT preserves both magnitude and phase information, enabling richer frequency representations, whereas DCT loses phase cues that are crucial for maintaining structural consistency.

Table~\ref{tab:ablation2}(b) evaluates the proposed GACA module against the standard cross-attention(CA). Replacing CA with GACA consistently improves accuracy, indicating that modeling local geometric relationships further enhances the model’s multimodal reasoning capability.

\subsection{Qualitative Analysis}
To evaluate MedFG-VQA against mainstream models, we conducted comparative experiments across multi-organ, multi-modal medical datasets. As shown in Figure \ref{fig:qualitative}(a), for lung X-ray questions, Gemma3 provides lengthy, detailed descriptions that include redundant information and overly long reasoning chains, with recommendations like ``further examination is required" that may lead to overdiagnosis. In contrast, our model delivers concise, accurate answers, directly stating ``no additional imaging or intervention is required". In Figure \ref{fig:qualitative}(b), for knee X-rays, Qwen3-VL describes multiple anatomical structures, some outside the visible range, while our model focuses on clearly visible key structures, such as the joint line, femoral and tibial articular surfaces, and joint space, with a succinct and targeted phrasing. In Figure \ref{fig:qualitative}(c), for a typical extra-axial mass on the dura mater, our model correctly identifies the lesion, while LLaVA-Med provides only a broad, non-specific description, missing critical diagnostic information. Overall, these examples demonstrate our model’s greater accuracy, clarity, and clinical relevance in medical visual question answering.

\section{Conclusion}
\label{sec:conclusion}

We present MedFG-VQA, a lightweight medical visual question answering (VQA) approach developed to overcome the dual challenges of data scarcity and computational constraints in clinical settings. Our model integrates global frequency-domain priors through FreqMemoryFusion (FMF) and uses a Graph-Aware Cross-Attention (GACA) mechanism to align visual-textual features while aggregating local structural information. This compact architecture delivers strong performance, demonstrating the feasibility of efficient vision language models for practical medical applications. To further facilitate research and development, we also constructed SynMedVQA, a large-scale medical VQA dataset with 2.059 million samples.

\noindent\textbf{Limitation.} The quality and diversity of the generated Q\&A pairs are inherently constrained by the capability boundaries of the underlying model. This dependency may hinder the framework's adaptability to emerging or novel imaging patterns that extend beyond the model's learned representations. Besides, our dataset is constructed based on single-view images, whereas practical clinical scenarios often involve multi-view and multi-modal imaging data, requiring joint reasoning and cross-modal integration for accurate diagnosis and assessment.

\small{\noindent\textbf{Acknowledgement.}
This work was supported by the National Natural Science Foundation of China (No. 62472222, U25A20442, 62427808), Natural Science Foundation of Jiangsu Province (No. BK20240080)

{
    \small
    \bibliographystyle{ieeenat_fullname}
    \bibliography{main}
}

\end{document}